# Are You Misinformed?
# A Study of Covid-Related Fake News in Bengali on Facebook

Protik Bose Pranto*
BUET
Bangladesh
1505044.pbp@ugrad.cse.buet.ac.bd

Syed Zami-Ul-Haque Navid*
BUET
Bangladesh
1505056.szuhn@ugrad.cse.buet.ac.bd

Protik Dey*
BUET
Bangladesh
1505051.pd@ugrad.cse.buet.ac.bd

Gias Uddin
University of Calgary
Canada
gias.uddin@ucalgary.ca

Anindya Iqbal
BUET
Bangladesh
anindya@cse.buet.ac.bd

## ABSTRACT
Our opinions and views of life can be shaped by how we perceive the opinions of others on social media like Facebook. This dependence has increased during COVID-19 periods when we have fewer means to connect with others. However, fake news related to COVID-19 has become a significant problem on Facebook. Bengali is the seventh most spoken language worldwide, yet we are aware of no previous research that studied the prevalence of COVID-19 related fake news in Bengali on Facebook. In this paper, we develop machine learning models to detect fake news in Bengali automatically. The best performing model is BERT, with an F1-score of 0.97. We apply BERT on all Facebook Bengali posts related to COVID-19. We find 10 topics in the COVID-19 Bengali fake news grouped into three categories: System (e.g., medical system), belief (e.g., religious rituals), and social (e.g., scientific awareness).

## KEYWORDS
Machine Learning, Social Media, Language, Crowdsourced, Dataset, Empirical study that tells us about people, Meta-Analysis, Content Analysis, Contextual Inquiry, Qualitative Methods, Quantitative Methods

## 1 INTRODUCTION

Nowadays social media sites such as Facebook, Twitter, etc. have become the most effective instrument of spreading news. Events occurring around the world reach us within minutes through news or articles published online and disseminated through social media. Unfortunately, the increasing popularity of these sites also invokes the possibility of spreading misinformation or fake news [51]. Compared to reputed media, the lack of editorial review and cross - checking of the credibility of an information cause a serious risk by numerous online media and posts of social media. Users in social media may fail to understand the unreliability of sensational news or analysis leading to wrong choice in their personal/social/ professional life. Fake news (i.e., misinformation) aims to deliberately propagate false information or hoax under the guise (e.g., in the context of a real news). In other words, misinformation appears as a legitimate piece of news but its integrity is tampered with by the content creator [58]. Fake news can create chaos and confusion in society and can damage the reputation of people, groups, or organization targeted by such misinformation [51].

As we have become more confined within the limited space of our home during COVID-19, people are relying more on Facebook during COVID-19 periods for getting news and information than other social media platforms or any other forms of news [36]. In these distressing times, people are more susceptible to this false news and are spreading them without knowing. Any misinformation regarding Covid-19 can have a disastrous effect on society. In 2020, almost half the fake news related to Covid-19 was propagated through social media like Facebook [67]. Therefore, filtering out of the fake news on Facebook is crucial. A lot of work are being done in detecting these Covid-19 related fake news. Most of these works are based on deep learning models [28, 54, 55, 91, 116, 117]. Fusing the language model, RoBERTa and the domain specific model, CT-BERT in one multiple layer perception achieved about 99.02% weighted average F1 score [20]. However, these works mainly focused on the detection of fake news in English posts.

Bengali is one of the most spoken and ancient languages in the world. According to www.berlitz.com, Bengali is ranked seventh in the list of most spoken languages with about 265 million speakers. Bengali is the mother-tongue of the people of Bangladesh. It is also the official language of West Bengal in India. Hence, the significance of Bengali is considerable and a large portion of the nation is quite active in social media. However, a good portion of Bengali social media users are not highly educated yet and often get influenced by media propaganda. Hence, the possibility of them being exposed to fake news is also growing at an alarming rate. For example, in Bangladesh, the price of salt was doubled due to a rumor of its shortage in 2019 [97]. The incident of Ramu in 2012 is also a tragic example of what fake information can do [15]. During COVID-19 periods, it was falsely reported that only 200,000 vaccines will be imported for the first installment and all of them will be given to government officials and ruling party members in Bangladesh [2]. In India, a health worker was arrested for spreading fake news about the number of Corona patients and the number of isolation wards setup for treatment [3]. These are only a few examples from a huge amount of false information circulating in the social media platforms and their consequences. Unfortunately, none of the currently available fake news detection tools have focused on Bengali posts on Facebook. Recently a multilingual model

---
*Equal Contribution

'FakeCovid' [86] is developed that can detect fake news in 40 different languages with an F1-score of 0.76. Fakecovid does not support Bengali. Moreover, the limited success of multilingual models like FakeCovid [86] shows that we need dedicated tools to detect fake news in Bengali.

In this paper, we have two goals: (1) Development of techniques to automatically detect COVID-19 related fake news in Bengali on Facebook posts, and (2) Empirical analysis of the detected COVID-19 related fake news posts in Bengali on Facebook. We answer four research questions (RQ).

**RQ$_1$ How accurately can Covid-19-related misinformation be detected in Bengali Facebook posts? (see Section 3)** Social media platforms, especially Facebook, are proving to be a viable source of recent news and information to the general public. It has become even more important to properly detect the Covid-19 related fake news that circulates through Facebook. First, we created a benchmark dataset of fake Bengali Facebook posts regarding Covid-19 using CrowdTangle [100]. CrowdTangle is a tool developed by Facebook to share Facebook posts and other social media aspects with researchers. Our benchmark dataset has 3,187 news of which 601 are fake. Second, we used three Transformer-based models to assess their performances. Our studied models include BERT [30], XLM-RoBERTa [27] and DistilBERT [83]. We find that BERT is the best performer with an F1-score of 0.97.

**RQ$_2$ How prevalent is Covid-related misinformation in Bengali on Facebook? (Section 4)** Covid-19 is currently the most discussed topic in the world. On the other hand, Bengali, as a language, has a special place in the world. So it is imperative to identify, collect and analyze the Bengali Facebook fake posts about Covid-19. For this purpose, we fetched all the Covid-19 related Bengali posts in Facebook from March 2020 to June 2021. Then we applied best performing model, BERT to filter out all the fake news. We analyzed these false news based on the demographics and and time and source of creation. We found that recently created Bangladeshi Facebook pages are more likely to spread misinformation. We also discovered that people are getting less interested in fake posts day by day.

**RQ$_3$ What topics are discussed in those misinformed posts? (Section 5)** To understand the fake Bengali news posted in Facebook, it is necessary to find out the types of false information are discussed in these posts. We applied topic modeling algorithm Latent Dirichlet Allocation (LDA) [18] on all the Facebook Bengali posts labeled as fake by our BERT model. We analyzed the news and categorized them according to the topic they were referring to. We grouped the news into ten different topics based on top words and top posts related to each topic. These ten topics are then grouped together into three categories: System (e.g., medical system), belief (e.g., religious rituals), and social (e.g., scientific awareness). The topics related to religious rituals and vaccination are the most discussed.

**RQ$_4$ How do people react across different topics of fake news? (Section 6)** To acquire a deeper perspective about how these fake news are affecting the general people, it is necessary to investigate their reaction to different topics of these fake news. We divided the available reactions of Facebook except *Like* into three categories: (i) Positive, (ii) Negative, and (iii) Haha. Then we explored people's reaction to different news topics by calculating the aforementioned reactions. We observed that political fake news have more negative reactions than any other topics. We also conducted demographic analysis of the frequency of the posts with respect to topics. We found that majority of the posts relate to vaccination and religious belief. We then manually examined the top 40-50 posts of each topics and associated comments. These comments reveal some interesting facts about the human nature. In most posts, people debate with the post's reasoning in both positive and negative ways. However, in religious-related posts, the majority of individuals avoid discussion. Rather than that, they frequently write the same comment (e.g., 'all praise to the almighty').

Our development of fake news detectors for Bengali posts on Facebook and the empirical analysis of misinformation of fake Bengali social media contents on Facebook can benefit several related stakeholders (see Section 7): health communities and religious institutions related to Bengali speaking communities, governments (e.g., Bangladeshi and Indian) and international communities, and above all, the general Bengali speaking population. The institutions and governments can learn about the different types of fake news found by our topic modeling, which can help them devise policies to mitigate the spreading of fake news. Tools can be developed based on our developed model to automatically warn Facebook Bengali users of the presence of fake news. The government/official authorities can also use our developed model to stay aware of the the topics related to fake news in social media.

**Replication Package.** Our study code and data are available at https://bit.ly/3l91YgW

## 2 BACKGROUND AND RELATED WORK

In this section, we first give a brief overview of the topic modeling algorithm that we used in our study (see Section 2.1). We discuss related work from the literature by dividing those into two categories: (1) Studies related to topic modeling in social media data (Section 2.2), and (2) Automatic detection and analysis of fake news in social media (Section 2.3).

### 2.1 Topic Modeling

In this paper, we use topic modeling to discover the types of Covid-19 related fake news in Bengali that are propagating through Facebook. We apply Latent Dirichlet Allocation (LDA) to determine to which topic each of the fake news belong. LDA is the most sophisticated and widely used method to obtain topics from textual data. Since topic modeling was first introduced, it has fascinated the researchers of various fields around the world. It has been vastly used in different branches of computer science such as bioinformatics [59], software engineering [6], linguistics, etc. Other than computer science, topic modeling is also used in other areas of science such as social science [75], nueroscience [24], etc. It has been applied frequently in the field of natural language processing as well [9], for example topic modeling has been used for classifying report data [57, 84], identifying topics in large amount of policy suggested by citizens [44].

Table 1: Words that must be present

| Keywords | Translation | Rationale |
|---|---|---|
| করোনা | Corona | There are various ways of alluding that the posts contain the word *coronavirus* in some fashion. |
| কোভিড -১৯ | Covid-19 | |
| কোভিড | Covid | |
| করোনাভাইরাস | Coronavirus | |

A topic model is a sort of plausibility inventive model which has the principal purpose of mining and exploring information [59]. It is mainly used to infer the underlying correlation between the words in textual documents [105]. Its goal is to find a cluster or group of words that are related with each other for interpreting the concealed relationship of a document [99]. The origin of topic modeling can be traced back to 1990 when Latent Semantic Indexing (LSI) was proposed by Deerwester et al. [29]. However, as it was not a probabilistic model, it could not achieve the desired quality. A lot of work has been done since then based on LSI [16, 37, 77]. Later in 2001, Probabilistic Latent Semantic Analysis (PLSA) was introduced which was the first genuine topic model [49]. Afterwards, Latent Dirichlet allocation (LDA) was proposed in 2003, which is the most complete and accurate topic modeling technique till date that uses probabilistic generative approach [18]. Since then all the progress that has been done in topic modeling are based on LDA.

The major use of topic modeling is in the field of natural language processing. In other words, it is mostly applied on textual data. From textual documents, it is quite easy and simple to understand the general idea of the text through information retrieval, but sometimes users need to understand the high-level themes of a corpus and explore the underlying patterns and dependencies [52]. Topic modeling offers to to paint a complete picture of this hidden relationships of the documents. It has been proved useful not only on large documents [106], but also on shorter texts [21].

## 2.2 Topic Modeling in Social Media

The emergence of social media platforms have taken the world by storm. In the early days of social media, people only discussed about their day to day life. But in recent years, social media platforms have become the center of all types of discussion, opinion and comments. Since then the subjects discussed in these platforms have been an area of interest for many researchers. In 2016, a practical topic model based on LDA was developed to obtain topics from social media corpus [80]. This study used a dataset of 90,527 textual document in the scope of aviation and airport management. Nowadays, Facebook is the most popular and widely used social media platform in the world. Thus analysing the contents of Facebook posts is a significant area of research. The posts and comments posted in Facebook have been analysed according to gender, topic and audience response [108]. Moreover these contents have also been experimented from academic [25, 115], medical [34], and political [73] perspective. Twitter is also a popular social media platform and empirical studies have been performed on modeling the topics of Twitter [11, 50, 113].

## 2.3 Fake News Detection and Analysis in Social Media

Beside print media, social media platforms have also become a major source of news and information for general people [48, 68]. People now can express their opinion and comment freely through these platforms. Around 35% Americans get news from Facebook only [88]. In Facebook, many groups, pages and individual post their belief on recent happenings of the world. This liberty to speak freely also poses the threat of fabricating fake news. As there is no inspection regarding the authenticity of news published in Facebook, people are being exposed to these misinformation which eventually affects the society [103].

Acknowledging the importance of detecting disinformation in social media, many works have been done trying to detect these types of false information effectively [28, 116, 117]. Data repositories containing the fake news in social media are created [51, 90]. Deploying data mining approach and deep learning methods are the most proven and efficient way to mark the fake news [54, 55, 91]. These deep learning methods not only use supervised learning [71, 79], but also make use of unsupervised learning [112]. Moreover, hybrid approach constituting both human-based and machine-based detection [70], geometric deep learning [64], Graph-aware Co-Attention Networks (GCAN) [62], graph neural networks with continual learning [46], etc. are also used. Besides news content, user profiles and social context can also be important factors for detecting these false information [92–94].

Besides machine learning approaches, the social media platforms also take necessary steps to fight fake news. Facebook and Twitter continuously remove accounts, pages, groups that are involved in fabricating false information [82]. Users can also be helpful in detecting all kinds of misinformation they encounter with in social media. They can report any post, tweet, comment that may seem misleading to them. There are also tools and extensions developed to understand and collect fake news [41, 87, 89]

There are also many surveys conducted to depict the human reaction regarding how they filter out the fake news from the authentic ones [65, 74]. People determine the authenticity of a post based on its source, content, and sometimes their predetermined notion about the news [36]. People were also asked to speak their minds while they were investigating fake news [39].

The fake news circulated by the popular social media platforms have a deep impact on the lives of the common people and therefore the society. The study of this influence has been of great interest to the researchers. For example, the fake news that were propagated through the social media during the 2016 US election is an incident that has been vastly studied [10, 107]. Social media created misinformation has been investigated from various perspective such as financial [31], medical [67, 95, 109], demographic [22, 66, 76], academic [33], etc. Covid-19 is the most talked about phenomenon in the world right now. Therefore, fake news regarding this deadly virus has become quite common. Many studies have been carried out on how Covid-19 related fake news spread on the social media and their impact on the lives of the general people [7, 23, 102].

## 3 HOW ACCURATELY CAN COVID-19 RELATED MISINFORMATION BE DETECTED IN BENGALI FACEBOOK POSTS? (RQ1)

### 3.1 Motivation

Fake news has indeed turned into a significant source of apprehension for the entire world. Since Bengali is 7th in terms of the number of speakers, detecting fake Bengali posts on Facebook is of paramount importance. The sheer volume of Bengali Facebook posts makes it a non-trivial task. Therefore, we need an automated system to carry out the detection. Numerous works have been done with regards to the detection of fake social media and news contents [20, 54, 55, 64, 86]. None of these models and datasets address the task for Bengali language. Although one work offered a dataset in Bengali as well as an NLP-based automated detector, they neither considered Covid-related misinformation nor did they collect posts from Facebook [51]. On this account, we need a detector that will be able to discern Covid-related Bengali fake posts on Facebook.

### 3.2 Detection Methodology Overview

Figure 1 depicts an overview of our methodology. First, we create a benchmark by collecting Bengali posts from Facebook and by manually labeling those (see Section 3.3). Second, we build several advanced language-based transformer models using the benchmark (Section 3.4). Third, we compute the performance of the models on the benchmark using 10-fold cross validation (Section 3.5). Fourth, we pick the best performing model based on the validation.

### 3.3 Benchmark Creation

Figure 2 shows the major steps we followed to create our benchmark dataset for fake news in Bengali. First we collect data from Facebook using a set of keywords. Second, we filter data and sample a subset of the filtered data for our manual labeling. Third, the first three authors together label the sampled posts for fake news. Fourth, we extend the benchmark using principles of active learning. Fifth, we finalize the benchmark by doing further rounds of labeling of the fake/true news posts created via the active learning. We discuss the steps below.

*3.3.1 Fetching Data with CrowdTangle.* To collect an inordinate quantity of data regarding coronavirus from Facebook we required an efficient and effective method. Therefore, we decided to use CrowdTangle, a Facebook owned tool [100]. CrowdTangle allows an authorized user to track, analyze and report on current trends and the impacts of those trends in the social media space. CrowdTangle only tracks public contents. In other words, posts on publicly available groups and pages, posts from verified public profiles are stored in the database of CrowdTangle. By fetching a certain post from CrowdTangle, a user can gain access to information such as when the post was posted, how the post is receiving attention (how many reactions, shares), which other public profiles or pages shared the post, etc. However, CrowdTangle never shares private information; for instance, the identity of the persons who reacted or commented on the post or any private post of any kind.

To obtain posts from Facebook that contain information in Bengali regarding coronavirus, we had to make use of certain keywords to deploy an effective search. We discussed among ourselves drawing on our experience to find those keywords. We decided to use the keywords showed in the Table 1 to make sure the posts are indeed about coronavirus. These are the words that must be present in the posts.

The data fetched using CrowdTangle did not include entire posts, rather truncated versions of them. As a consequence, we used Trafilatura, a web scraping tool, to follow the URLs of the posts provided by CrowdTangle to obtain entire texts.

*3.3.2 Data Filtering and Manual Labeling.* We divided our dataset building into two phases. In the first phase we manually labeled and filtered data. The first phase will be described in the rest of this subsection.

We used CrowdTangle to fetch statuses and shared links posted between March and June in 2020. We decided to forgo videos and texts because we are only dealing with textual representation of information in our study. We began investigating posts starting from March due to the fact that the first covid-infected patient in Bangladesh was discovered on 8th March, 2020 [111].

In the first phase, we manually perused 48000 posts fetched by CrowdTangle between March and June of 2020. We kept any relevant post in our dataset except the ones that meet the following criteria.

(1) **Duplicate Posts:** We discarded any post that we had already seen before.
(2) **Undisputed Faith:** We disregarded any post that preached faith and did not conflict with any scientific fact or proven truth. It is not our purpose to disparage or promote any faith in anyway. Hence, we left these posts of our dataset so that we do not introduce bias in the studied deep learning models.
(3) **Irrelevant Posts:** We ignored any post that contained the desired keywords but did not offer any context regarding the pandemic.

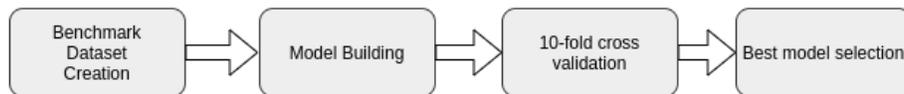

Figure 1: Best Model Selection Flowchart

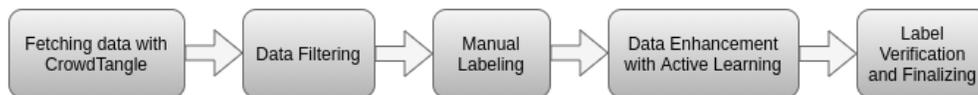

Figure 2: Benchmark Dataset Creation Overview

(4) **Unverifiable Posts:** We also left out the posts of which we failed to verify either the legitimacy or the falsehood against any authentic data source.
(5) **Local Death or Infection Statistics:** We discarded news that only offer number of deaths by Covid or infected patients in the local area. Since these pieces of news offer no context or additional information about Covid except sheer numbers, we did not include them.

After the rigorous filtering process, we managed to collect 676 unique posts. Among those 676 posts, 513 were labeled true, while 163 were labeled false. To label each post, we extensively cross-checked all the references and information in that post. The posts that conveyed sarcasm or a travesty of the information related to Covid were labeled as fake news. To ascertain the authenticity or falsehood of the rest of the posts, we relied on trustworthy data sources. For instance, reputed national newspapers such as Prothom Alo, Daily Star, Dhaka Tribune, international media such as BBC World, Reuters, Times of India and health organizations such as World Health Organization, Centers for Disease Control and Prevention etc. A quintessential example of such verification would be the following myth about Covid diagnosis that had been circulating in the social media space.

> *Diagnose yourself for corona virus (Covid-19): The test is: Get up every morning, take a deep breath in a clean environment and hold it for a little over ten seconds. If you do not have a cough, do not feel chest pain or pressure, do not feel any discomfort while holding this breath, it means that there is no fibrosis in your lungs, that is, there is no infection, you are completely healthy.*

The aforementioned fake news has been debunked in several reliable media sources. We affirmed the falsehood of this news from Times of India [69].

*3.3.3 Dataset Enhancement with Active Learning.* After collecting 676 data points, we adopted a process of enlarging our dataset that would be less time and effort consuming. We employed active learning [85] to facilitate data collection in a more efficient manner. The idea behind active learning was to have a learning algorithm learnt from fewer labeled data [85]. An active learner can distinguish between what is worth learning and what is not. To put simply, an active learner can choose the training data it learns from [85].

Our dataset enhancement with active learning works in the following steps:

(1) First we trained our active learning model on the 676 data points we collected and manually labeled.
(2) Then we fed the model with unlabeled data
(3) The model labeled a post as true if the confidence score was 97% or higher. Conversely, the model labeled a post as false if the confidence score was lower than 35%.
(4) We manually verified all the labeled data found in the previous step as described in the Section 3.3.2.
(5) We inserted the data that the model managed to correctly label into our dataset. Thus, we had a larger dataset.
(6) We trained the model on our enlarged dataset to bolster the learning process.
(7) Go back to step 2

We used pre-trained multilingual DistilBERT to carry out active learning. We used a simple sigmoid layer on top of the pre-trained checkpoint to compute the confidence score. Finally, we trained the model for 40 epochs with a mini-batch size of 8 and thus fine-tuned the model on our dataset. DistilBERT is a relatively smaller variation of BERT in terms of number of parameters. While DistilBERT is 40% smaller than BERT, it retains 97% of its performance [83]. The simultaneous learning and labeling process is depicted in the Figure 3.

Having applied active learning, we verified the labels returned by the model manually. Thus, we managed to collect 2511 data points in addition to our previously manually collected 676 posts. Of the 2511 posts, 2073 were labeled as true while 438 were false. Therefore, our dataset had 3187 posts in total, of which 601 posts were false or fake. Table 2 summarizes our dataset.

*3.3.4 Label Verification and Finalizing.* Active Learning is performed incrementally. In each iteration, we applied the DistilBERT model on unlabeled data and the model returned predicted labels (as explained in Section 3.3.3). We validated the model's predictions by cross-checking against verified news sources. However, posts that conveyed sarcasm are labeled as Fake (or False). We also discarded irrelevant posts following the same criteria as mentioned in Section 3.3.2. Correctly labeled posts are added henceforth to the already labeled dataset and thus enhancing it. We applied the same labeling and filtering process as described in Section 3.3.2 during this enhancement procedure.

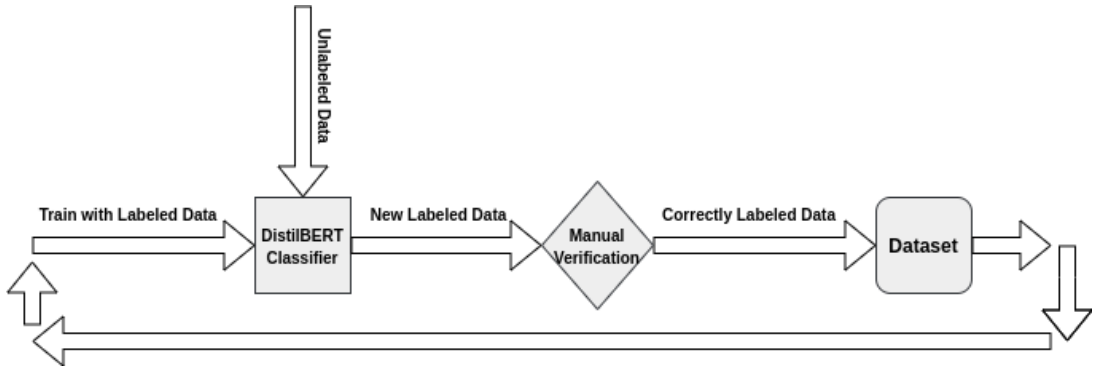

Figure 3: Data enhancement with Active Learning

Table 2: Summary of Benchmark Dataset

|  | Manual Collection | Active Learning | Total |
|---|---|---|---|
| True | 513 | 2,073 | 2,586 |
| False | 163 | 438 | 6,01 |
| **Total** | 676 | 2,511 | 3,187 |

### 3.4 Model Development

*3.4.1 Model Overview.* We used three Transformer-based pre-trained models to perform experiments on our dataset. These models, being pre-trained, perform considerably better than shallow neural networks such as LSTM, Bi-LSTM on dataset that is on the relatively smaller side such as ours [55]. Furthermore, these models learn contextual representations, allowing them to attain state-of-the-art performance in text-classification tasks [63]. Firstly, we delineate the models we experimented with in our study before describing the experimental setups we adopted for each model.

(1) **BERT:** BERT, which stands for Bidirectional Encoder Representations from Transformers [30], is a bidirectional context encoder model that is built on top of the Transformers [104]. This model was pre-trained by conflating two objectives: masked language modeling and next sequence prediction. Since, we had access to only 12 GB of RAM provided by Google Colab, we chose BERT-base instead of BERT-large to address the memory constraints. Furthermore, we used the bert-base-multilingual-cased checkpoint for our experiments. This particular checkpoint was pre-trained on a huge corpus comprising texts in top 104 languages in Wikipedia.

(2) **XLM-RoBERTa:** This is a Transformer-based multilingual model that employs masked language techniques [27]. This model has been pre-trained with efficient training methods similar to RoBERTa introduced by Facebook [60]. Unlike BERT, it does not use language embeddings and hence is capable of identifying the language by itself. In our research we used the xlm-roberta-base that was trained on more than 100 languages.

(3) **DistilBERT:** DistilBERT is a significantly smaller variation of BERT in terms of number of network parameters [83]. It runs 60% faster than BERT while retaining 97% of its performance with regards to language understanding. We made use of the distilbert-base-multilingual-cased, a multilingual checkpoint of the model. It was trained on the same corpus as BERT.

All the models we have experimented with have similar architecture. Table 3 summarizes the technical intricacies of each model. Here, L, H, and A stand for number of layers, hidden size, and number of self attention heads, respectively [27, 30].

*3.4.2 Experimental Settings.* All of the models above employ bidirectional Transformers. Consequently, they are able to capture context from both right and left. However, the deep bidirectional representations they return from being pre-trained on unlabelled sentences alone is not enough to classify texts. Therefore, we used the versions of the models that are capable of downstream tasks such as classification. These classification models simply attach a classification head (a linear layer) on top of the raw models (BERT, XLM-RoBERTa and DistilBERT). As for model optimizer, we used AdamW [61]. Moreover, we utilized cross-entropy losses to facilitate the learning procedure.

The general steps of the fake news detection process is depicted in the Figure 4. The preprocessed input is passed to the Tokenizers of the corresponding models, such as BertTokenizerFast, XLM-RobertaTokenizerFast etc. The tokens from the Tokenizers are then passed to the models which return deep representations of the input texts. The deep representation is a 768 dimensional vector. Finally, the classification head returns probabilities for each label.

Table 3: Summary of Model Architecture

| Model | Checkpoint | L | H | A | Total Parameters |
|---|---|---|---|---|---|
| BERT | bert-base-multilingual-cased | 12 | 768 | 12 | 110$M$ |
| XLM-RoBERTa | xlm-roberta-base | 12 | 768 | 12 | 270$M$ |
| DistilBERT | distilbert-base-multilingual-cased | 6 | 768 | 12 | 66$M$ |

We used the simpletransformers implementation for the models, which in turn uses the reputed opensource Transformers library huggingface. We performed experiments on NVIDIA Tesla K80 GPU provided by Google Colab.

*3.4.3 metrics.* In our data *True News* is represented as *Positive* class, while *Fake News* is labeled as *Negative* class. Consequently, TP (True Positive) denotes the discovery of a true news as opposed to TN (True Negative) reporting fake news. The definition of FP (False Positive) and FN (False Negative) with regards to our dataset follows accordingly.

We report performance using three standard metrics of information retrieval: precision (*P*), recall (*R*), and F1-score (*F*1).

$$P = \frac{TP}{TP + FP}, \quad R = \frac{TP}{TP + FN}, \quad F1 = 2 * \frac{P * R}{P + R}$$

$TP$ = # of true positives, $FN$ = # of false negatives, $TN$ = # of true negatives, and $FP$ =# of false positives. Here, precision is the proportion of actual true news with respect to the number of news that were labeled true by the models. Recall is the ratio of news that had been correctly labeled true and the total number of true news in the test dataset. Precision determines the efficacy of the models at accurately labeling any class. Conversely, Recall decides whether the models can completely represent each class. Since, these two metrics share an inverse relationship, F1-score, the harmonic mean of them can illustrate a more comprehensive picture of the performance. Furthermore, F1-score has been considered a preferred metric when the dataset in imbalanced, similar to our scenario [47].

In terms of performance measurement of deep learning models, Accuracy is an extensively used metric. $Accuracy = \frac{TP+TN}{TP+FP+TN+FN}$ However, given that our dataset is imbalanced (i.e., fewer fake news posts), Accuracy can prove to be a biased measurement of performance. The sheer volume of True Positive can give a misleading result [47]. Therefore, we report our results in terms of Precision, Recall, and F1-score.

*3.4.4 Hyperparameter Tuning.* In the following subsection, the hyperparameter tuning process through which we sought the best settings for each architecture is illustrated. To carry out the tuning procedure, we took inspiration from state-of-the-art text classification literature [30, 42, 45]. For optimal performance, we fine-tuned the following parameters: (1) Initial learning rate, (2) Batch size, (3) Number of epochs, (4) Learning rate decay scheduler, (5) Power factor for polynomial schedule with warm-up, (6) Regularizer

We have employed the random search technique instead of the standard grid search for fine-tuning. Empirical study maintains that the former is more efficient and less resource demanding [17]. We have experimented with the following values for each of the hyperparameters: ($1e^{-5}$, $2e^{-5}$ and $4e^{-5}$) for learning rate, (15, 17 and 25) for number of epochs, (*linear with warm-up*, *constant*, *cosine with warm-up*, and *polynomial schedule with warm-up*) for initial learning rate decay schedule. Furthermore, we conducted experiments with two values (1 and 2) for the power factor of polynomial decay schedule. Due to the limited capacity of memory in our hardware setup, we carried out our investigation with rather smaller mini-batches (4 and 8). Additionally, we employed L2 regularization to each of the models. Incidentally, performance of BERT improved considerably after applying L2 penalty. We will discuss the said improvement in section 3.5. Table 4 highlights the best settings for each model.

In addition, we have left the other hyperparameters such as *adam-epsilon*, $\beta 1$ and $\beta 2$ to their default values ($1e^{-8}$, 0.9 and 0.999 respectively) [30, 98]. We have also allotted 512 tokens from each data point to be fed to the models [30].

> **Findings 1:** BERT tended to overfit on our dataset more than the rest of the models. Hence, regularization was applied on BERT to allow it to generalize more on unseen data. F1-score escalated from 0.94 to 0.97 after applying L2 regularization.

## 3.5 Model Performance

We experimented with three Transformer-based models enumerated in subsection 3.4.1 to discern how accurately we can detect fake news from our proposed dataset. To evaluate the performance of the models, we performed stratified k-fold cross validation. Stratified cross-validation mitigates the variance in the results of classifiers and gives stable estimation about the performance [32, 38]. This technique is conducive for deriving accurate measure of performance on a relatively small and imbalanced dataset such as ours.

In our experiments we set k to 10. Hence, we ran iterations of train and test cycles. In each iteration, the dataset was split into 10 folds, among which 9 folds were used for training, while the remaining fold was used for validation. In addition, the ratio of true and false samples was maintained in both train and test folds in each iteration. To report the performance metrics, we take the mean across 10 iterations.

Table 6 reports the performance of all the models we experimented with on our dataset. BERT happens to be the best performing model with F1-score at 0.97, closely followed by XLM-RoBERTa at 0.95. From the results, we see that XLM-Roberta has achieved a recall score almost as good as BERT. XLM-RoBERTA managed to detect true news reasonably well, which led to a lower FN and higher TP and thus, a high recall. On the other hand, the model has a considerably lower precision than BERT. XLM-RoBERTa has generated a noticeable number of FP which denotes the relative inefficacy of detecting false news. To summarize, XLM-RoBERTa became better at detecting true posts than false posts. Conversely, BERT proved to be equally good at detecting both false and true news as its precision and recall scores are almost the same.

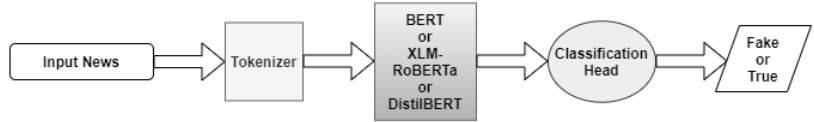

Figure 4: Classification steps

Table 4: Fine-tuned hyperparameters

| Model | # Epochs | Batch size | Learning Rate | Learning rate decay | L2 penalty |
|---|---|---|---|---|---|
| BERT | 15 | 8 | $1e^{-5}$ | cosine with warm-up | 1 |
| XLM-RoBERTa | 15 | 8 | $2e^{-5}$ | polynomial with warm-up (power=2) | 0 |
| DistilBERT | 15 | 8 | $1e^{-5}$ | cosine with warm-up | 0 |

Furthermore, we made another observation regarding the ability of XLM-RoBERTa and BERT models to generalize performance on validation data. Whereas the performance of XLM-RoBERTa was consistent across train and validation datasets or folds with L2 penalty set to 0, BERT tended to overfit on the train data. Consequently, there was a significant improvement in the performance of BERT after we assigned the L2 penalty to 1. Table 5 depicts the change in performance after applying regularization.

However DistilBERT, though scored over 0.9, performed rather poorly compared to BERT and XLM-RoBERTa. This is expected since the distillation process lowers the performance of DistilBERT significantly [83].

In the following, we inspect several posts that our best performing model, BERT, managed to classify correctly. Moreover, we take a look at several other posts that BERT labeled incorrectly. The posts have been translated into English. The following is an example of a *True Positive*.

> "Tough lockdown is the key to success, Corona-free New Zealand: Incidentally, the first corona outbreak occurred in New Zealand on February 28 this year. The last time a new infection was reported was on May 15. The total number of victims in this country of fifty million people was 1154 people. Twenty-two people have died. No infection has been reported in the last 18 days. The lockdown has been lifted from the country since midnight on Monday. Although there are strict restrictions at the border, there are no restrictions at present. The small country has benefited from three consecutive months of severe lockdowns and a large number of corona tests. It is to be noted that a seven-week lockdown was issued after the first corona-infected person was found. No one was allowed to leave the house."

The next excerpt is an instance of *True Negative*.

> "Corona does not spread at temperatures above 30 degrees, researchers claim. The researchers said the virus could not survive on objects or in the air at temperatures above 30 degrees. As a result, they think that tropical countries like Bangladesh are relatively safe. Experts are discouraging the use of Air Conditioners as well as ensuring cleanliness and safe distance to reap the benefits. Researchers are giving various suggestions to control the spread of Covid. That being said, high temperatures and high water vapor in the air prevent the virus from spreading. Researchers in Singapore say temperatures above 30 degrees Celsius and humidity above 70 per cent in the air prevent corona from spreading."

The following is an example of *False Positive*.

> "Professor Oliver Linton, a British researcher, gave good news to the world. He said the transmission of corona virus will decrease significantly in the next 40 days. The researcher, a professor of political economy at the famous Cambridge University in London and a fellow at Trinity College, said: 'We will start to observe the decline in the next 40 days.' According to him, this catastrophe called corona will be diminished and the number of new infections will be greatly reduced. Then a very small number of people will die because of it. Professor Linton said in his article that in the final days of the Corona virus, more than 60,000 people would be infected and more than 10,000 would die every day in the world. After that, the infection by corona will continue to decrease around the world. But before April 17, bad times would be over. This means that the virus has already reached the final stage of its infection."

The next post is a *False Negative*.

> "The French government has banned the use of controversial and 'harmful' Hydroxychloroquine as the treatment of patients with coronavirus. The country issued a ban this week after two French consulting firms and the World Health Organization warned against the drug. Hydroxychloroquine is commonly used as a medicine for Malaria, Arthritis or skin infections. While some

Table 5: Impact of regularization

| Score | L2 Regularization parameter = 0 | L2 Regularization parameter = 1 |
| --- | --- | --- |
| F1 | 0.94 | 0.97 |

Table 6: Performance of the models

| Model | Precision | Recall | F1-score |
| --- | --- | --- | --- |
| BERT | 0.96 | 0.97 | 0.97 |
| XLM-RoBERTa | 0.93 | 0.96 | 0.95 |
| DistilBERT | 0.92 | 0.91 | 0.91 |

*claim that the coronavirus drug is effective, many countries allow it to be used to combat the ongoing epidemic."*

> *Findings* 2*:* BERT stands out as the best-performing model with an F1-score of 0.97. XLM-RoBERTa was better at detecting true news than fake. BERT was equally good at recognizing both the classes of news.

## 4 HOW PREVALENT IS COVID-RELATED MISINFORMATION IN BENGALI ON FACEBOOK? (RQ2)

### 4.1 Motivation

Social media has evolved to become an integral part of our daily lives. With the rise of social media, particularly the Facebook News Feed, the prevalence of fake news has grown. *The International Federation of Library Associations and Institutions (IFLA)* highlighted several points to help users identify misleading news [1]. Often, these hoaxes are so compelling that it is impossible to tell fact from fiction. As one of the most widely spoken languages globally, Bengali posts are more likely than those in other languages to promote fake news. Currently, Covid is a worldwide trending topic. On Facebook, there is presently no survey on Bengali Covid-related fake news. We study the origins, characteristics, and distribution of these posts and the reactions of people from various countries to them. This, we hope, would assist a large number of individuals in deciphering misleading Bengali Facebook posts.

### 4.2 Approach

According to what we have already stated, Bengali is primarily spoken in Bangladesh and some regions of India. It is also said in a few other countries to a considerable extent where a large number of Bengali people have migrated. The United Kingdom, the United States, the Middle East, Canada, Australia, and some European countries have millions of Bengali-speaking first-generation migrants. To make it simple, we have divided all the countries into three groups - *Bangladesh*, *India*, and *Other*.

The purpose of this research question is to examine the prevalence of misinformation about Covid in Bengali Facebook posts from the standpoint of our three primary demographic classifications. We study the characteristics and sources of fake news and discovered the following absolute and relative impacts in each demographic group as follows.

**Fetch Data from CrowdTangle.** We have used *CrowdTangle* to retrieve Facebook posts once again, using the exact keywords mentioned in Table 1. This time, we extend the period starting from March 2020 to June 2021. In the preceding section, we considered two methods for labeling data: manually and through active learning. However, in this section, we use *BERT*, which performed better on our benchmark dataset, to classify the posts as true or false. Thus, we have accumulated 43979 posts, of which 6012, or roughly 13.67%, are fake. We used this newly created dataset to answer all the following research questions, including this.

**Study from Demographic Standpoint.** We separated fake news from our freshly obtained data and classified it into three groups based on the location of the post's origin. To determine the source of these posts, we checked the creation years of the Facebook pages responsible for propagating fake news and grouped them accordingly. Additionally, we examined the type of posts to determine how incorrect information is most frequently conveyed and how different individuals react on them.

Finally, we calculated the *Overperforming Score*. Overperforming score is a metric calculated by CrowdTangle for each post we fetched off the Facebook. Overperforming score can be expressed using the following equation.

$$Overperforming\ score = \frac{Total\ Interaction}{Expected\ interaction} \quad (1)$$

Expected interaction is computed for a page or a person. If the post in question is posted from a personal account, expected interaction is calculated for that person. Conversely, if it is a post on a public page, expected interaction is computed for that page. In either case, mean of the last 100 posts is considered as the expected interaction. Therefore, overperforming score is the ratio of actual total interaction and the expected interaction of a post under consideration. In other words, it is a measure of how well or how poorly the post had done in the eyes of the Facebook users. If a post receives fewer interactions than the expected interaction, the equation is inverted and negative of the ratio is taken. Thus, a positive overperforming score suggests that the post is doing rather well according to the people on social media, while negative overperforming score denotes the opposite.

### 4.3 Results

We analyzed Facebook pages and their posts containing false information to determine the traits of fake news and how people in various countries react to it.

Table 7: Percentile Frequencies of Types of Posts

| Types | In Bangladesh | In India | In Other Locations |
|---|---|---|---|
| Link | 60.38% | 74.90% | 70.79% |
| Status | 39.57% | 25.10% | 29.21% |
| Live Video | 0.05% | 0.00% | 0.00% |

**Origin of False Propagator Pages on Facebook.** On 4 February 2004, the facebook.com was launched for Harvard University students in Cambridge, Massachusetts. Harvard students were able to build a profile page with personal information and communicate using Facebook. These profile pages were updated in July 2008. Then, in addition to individual profile pages, new types of Facebook pages became publicly available. [19]. We executed a demographic analysis of the pages responsible for spreading fake news. According to the figure 5, between 2008 and 2021, the majority of pages in Bangladesh were created in recent years, even in 2021. This is true for other regions also. It appears that in general the pages created in the early ages are more reliable in these regions. Whereas in India, a large number of pages propagating misinformation were formed in 2009.

> *Findings* 3: Between 2008 and 2021, the majority of Bangladeshi Facebook pages that spread misinformation were created; this is also true for the rest of the world, except for India. Long-running pages appear to be more trustworthy.

**Characteristics of Different Types of Posts.** Here, we will discuss only three distinct sorts of posts. The first is *Link*, which involves a post with a news article or an external URL link. The second is *Status*, which is an updated feature that enables users to interact with their peers about their thoughts, whereabouts, and valuable information. Lastly, there is *Link Video*, that is a video associated with some captions. As observed in Table 7, the majority of fake news is coupled with a news link, which is applicable to all regions. On the other hand, video-related fake news are rarely recognized due to the risk of being exposed.

Whenever any new issue evolves, at first it becomes highly popular, but after a while, enthusiasm fades away. Meanwhile, another topic is raised, and the attention of people diverts towards it. For example, the first Covid case was found in March in Bangladesh, as previously stated. At the time, it was a popular topic in the country. There had been a lot of news moving around, and people were acting accordingly. At that time, the prevalence of fake news was fairly great than any other time. Naturally, reliable sources were relatively silent at that time as very little concrete or scientifically accepted information was available at that time. As the days passed, the frequency of these fraudulent news gradually decreased until a new concern arose. This is clearly visible from the figure 6 where we have showed that frequency of post creation in a month for *Bangladesh*, *India* and *Other*.

To maintain simplicity, we split the data into two halves. The first component is for 2020 and the second is for 2021. The subfigure 6a depicts three zigzag lines representing the percentage of post-creation activity in three distinct regions, which is likewise applicable for the second subfigure 6b.

> *Findings* 4: Across all regions, more than half of posts are written in text format rather than video or image. Since, videos are easier to verify, false information is generally transmitted via text.

**Responses of People from Different Countries.** People engage in communication on Facebook via three behaviors—*Reaction*, *Comment*, and *Share* [56]. Facebook reactions, released in February 2016, are an extension of the old "Like" button. Its six options (Like, Love, Haha, Wow, Sad and Angry) are represented by slightly edited versions of several long-established Unicode Emojis, and they allow for a more nuanced expression of how users feel towards a post[101]. The more reactions, comments, and shares a Facebook post receives, the more accessible it becomes to other users. Such a high level of user activity on a fake news site indicates that it has spread widely and will influence adversely on a large number of people. As per the table 8, people in India and other countries are more likely to comment on and share false news than individuals in Bangladesh. Whereas the users from Bangladesh tend to interact with such news via the Facebook responses options.

> *Findings* 5: When a new issue arises on social media, it quickly gains widespread attention. At times, the likelihood of fake news spreading is high. However, over a while, curiosity fades away, and the chance of spreading misinformation decreases. In March, Bangladesh discovered its first covid patient. It was the most discussed topic in this region at the time, and a significant amount of fake news propagated. However, as people's curiosity reduced, their reaction to false information declined as well.

Table 8: Average Interactions, Share and Comments on Bengali Fake Posts

|  | Bangladesh | India | Other |
|---|---|---|---|
| Avg. Interactions | **5029** | 3911 | 3264 |
| Avg. Comments | 175 | **235** | 147 |
| Avg. Shares | 537 | **689** | 363 |

**Comparison by Overperforming Score.** We calculated the mean and the standard deviation of the overperforming scores of all the true news and all the fake news separately. The results are summarized in Table 9.

From Table 9, we see that there is a considerable disparity between true and fake news with regards to standard deviation of overperforming score. We took the mean of overperforming score of fake news found each month within March, 2020 and June, 2021.

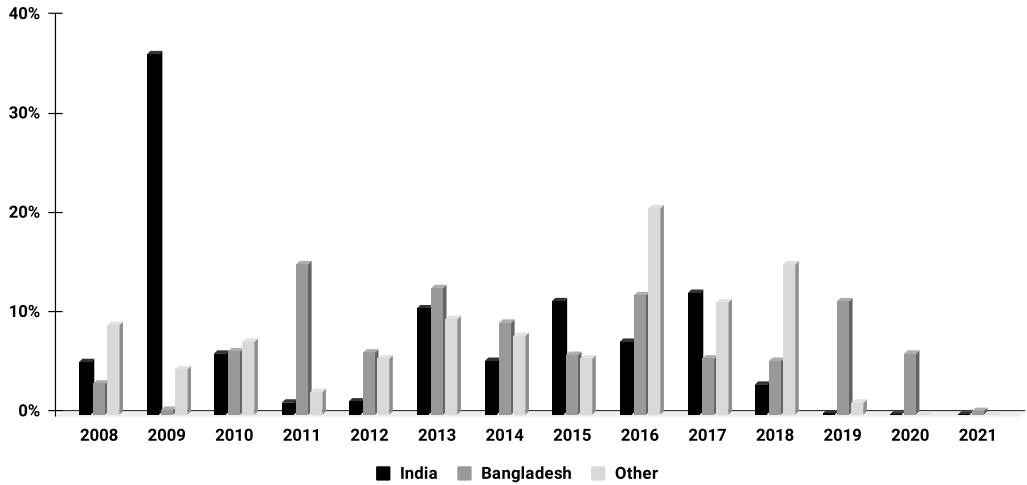

Figure 5: Percentile Frequencies of Page Creation (With Year)

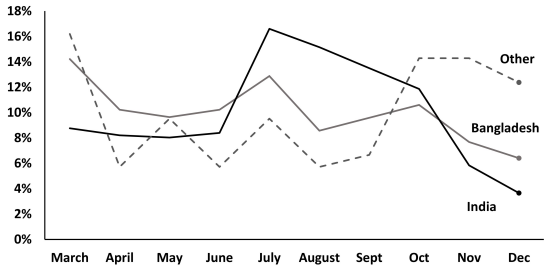

(a) March - December (2020)

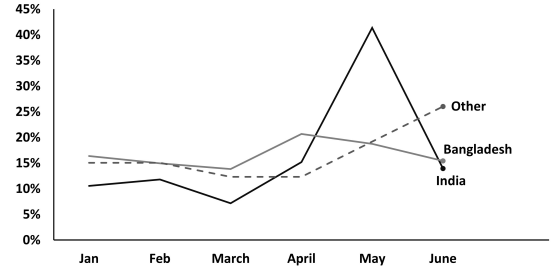

(b) January - June (2021)

Figure 6: Percentage of Posts Creation

Table 9: Comparison of True and Fake News in terms of Overperforming Score

| Label | Mean  | Standard Deviation |
|-------|-------|--------------------|
| True  | 15.36 | 67.86              |
| Fake  | 20.61 | 90.03              |

Figure 7 depicts the downward facing mean overperforming score of fake news over time. The steep decline of the curve suggests how people on social media became aware of misinformation and became increasingly immune to it as new information about coronavirus came to light.

## 5 WHAT TOPICS ARE DISCUSSED IN THOSE MISINFORMED POSTS? (RQ3)

### 5.1 Motivation

Misinformation has spread everywhere, from medical science to religion. The public is increasingly being misled on a variety of issues. Due to the nature of these claims, it can be somewhat difficult to verify them. Sometimes, individuals intentionally promote false narratives for political gain (US Election 2016) or undermine social cohesion in targeted communities. As a result, it is essential to understand the topics discussed in posts that include inaccurate information.

### 5.2 Approach

To identify a topic, we must first assign it a label that briefly expresses the topic's fundamental concepts. To solve this research question, we will conduct a two-step process of topic identification. We will perform *Topic Modeling* in the first stage to find an optimal number of topics as well as generate topics and then label the topics in the second step. Finally, we categorize all the topics into categories and subcategories.

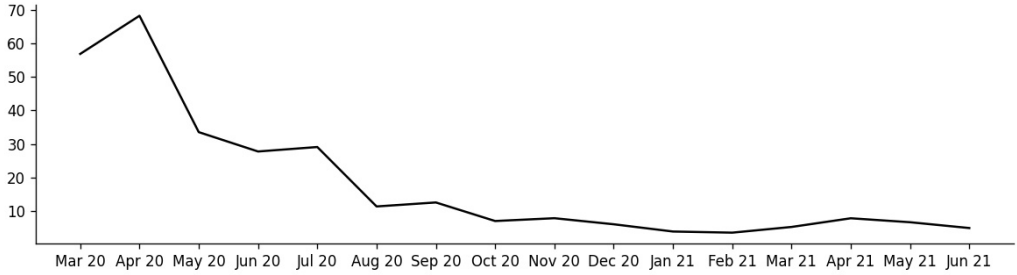

**Figure 7: Mean Overperforming Score of Fake News by Month**

**Step 1. Determine the optimal number of topics.** We use the *Latent Dirichlet Allocation (LDA)* technique from the *Gensim* [78] library to create topics. This algorithm takes a number of posts as input and generates a list of topics for grouping the posts into *K* topics. Arun et al. proposed a standard practice of determining the optimal topic count based on the *coherency* among the topics. The optimal number of topics will result in the maximum coherence value [13]. To calculate the coherence score, we deploy standard *c_v* metric which is available at *Gensim* library.

We apply LDA on our data for $K = [5, 8, 10, 12, 15, 17, 20, 23, 25, 30, 35, 40, 45, 50]$. We ran each model with 400 iteations and 40 epochs. We calculate the coherence score of the generated topic model for each *K*. We choose the topic model with the highest coherence score, which in this case is 0.468 for $K = 15$. Thus, this is our optimal topic number.

**Step 2. Generate topics.** The model provides the following information for each topic: A list of the top *N* words that best describe the topic, together with the probability of each word, which indicates the word's relative defining power for the topic and a list of the *M* posts that have been made about the topic. Correlation values between 0 and 1 are assigned to each connected post. The greater the correlation between two posts, the more relevant the post is. We categorize the topics based on the posts with the highest correlation values.

**Step 3. Label all topics.** Following prior work on topic labeling [4, 5, 8, 14, 43, 81, 114], we manually label each topic using an open card sorting approach [53]. There is no predefined label for a topic in this method, since such a label is determined throughout the open coding process. To label a topic, we rely on two types of information mentioned in the previous step: (i) The list of top words in the topic and (ii) A list of top 40 − 50 posts linked to the topic. Initially, three authors were involved in the process of labeling. Then the remaining two authors who individually have 15 years of experience in empirical software engineering and software design joined. All the authors discussed each topic and came up with a label for it. The coders iterated over each topic's labeling until consensus was attained.

Following this labeling, we integrated a number of topics that were similar but used separate vocabularies, which LDA classified as independent. For instance, we merged topics 13 and 14 into religious rituals because both topics featured texts on conducting religious practices relevant to the Covid circumstance. However, each religion has its own special words that is utilized only inside that religion. As a result, LDA divided those into two distinct topics. Finally, We obtained 10 different topics.

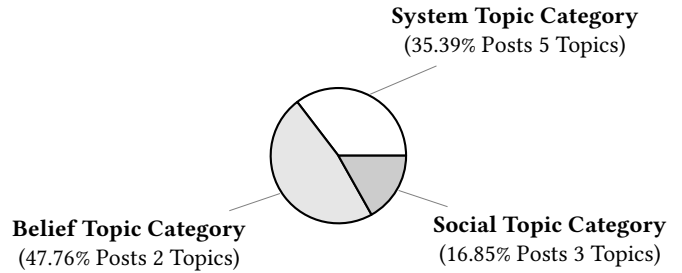

**Figure 8: Distribution of high-level topic categories of Bengali fake news on Facebook**

**Step 4. Categorize into higher categories.** We group all the topics into higher categories. For example, the two topics *Medical System* and *Vaccination* are related to *Medication System*. As a result, we've classified the two topics as *Medicine*. This process was repeated number of times to develop even higher categories, until no further higher categories could be obtained.

### 5.3 Results

We found a total of 10 topics. After labeling the topics, we grouped them into three high-level categories: **System**, **Belief**, and **Social**. Figure 8 shows the distribution of topics and questions in the three categories. Among the categories, *Belief* has the highest coverage of topics (47.76% of posts), followed by *System* (33.3% of posts), and *Social* (16.86% of posts).

Figure 9 illustrates the 10 topics in order of the number of posts in percentile. The topics are arranged in ascending order, with the most significant portion on the left. Out of all topics, posts related to *Religious Rituals* are found majority of the time (44.27%), followed by *Vaccination* (23.74%), *Scientific Awareness* (13.69%), and so on.

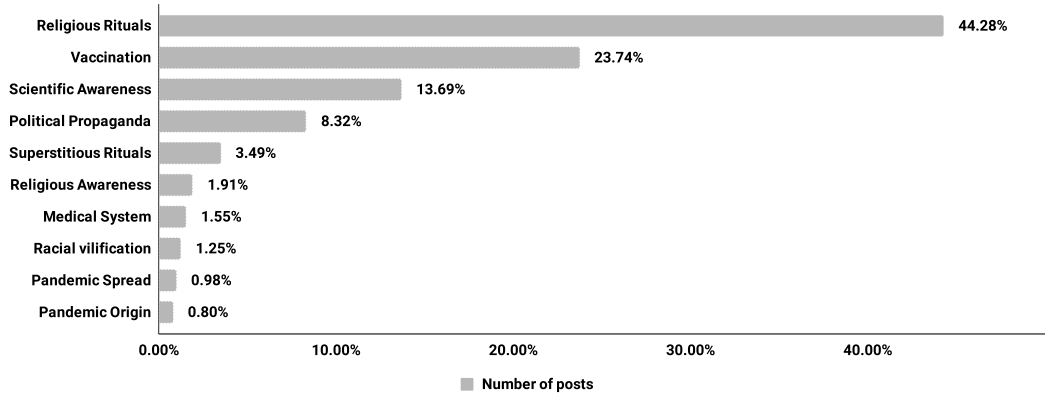

Figure 9: Distribution of topics related to Bengali fake news on Facebook based on percentage of posts

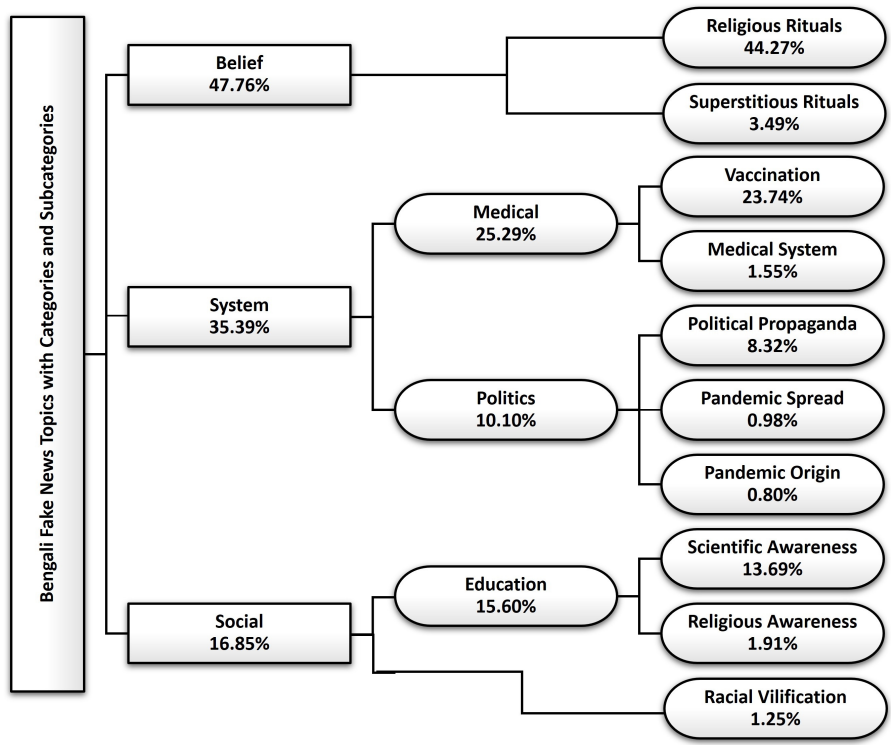

Figure 10: Topics with categories and subcategories

Figure 10 shows the 10 topics grouped into three categories. We subdivide each high-level topic category into sub-categories. For instance, *Social* category has two sub-categories: 1. *Race* 2. *Education*. Each sub-category is divided into a number of topics. For example, there are two topics under the sub-category *Education* - 1. *Scientific Awareness* 2. *Religious Awareness*. Finally, in such way, the topics of all the sub-categories combine to become the final 10 topics.

We now discuss the topics by three categories below:

**Belief Related Topics.** This is the broadest category of the three. This category has two topics: *Religious Rituals* (44.27%) and *Superstitious Rituals* (3.49%). Both are connected to human belief. *Religious Rituals* are primarily concerned with religion-based rituals to eliminate Covid. On the other hand, *Superstitious rituals* are any belief or behavior that is considered irrational, supernatural, or fearful of the unknown by non-practitioners.

**System Related Topics.** Five of the ten topics fall into this category. These topics include vaccination, the medical system, and internal and international political issues. This category is subdivided into two subcategories: (*i*) *Medical* (25.29%), (*ii*) *Politics* (10.10%). The former holds the discussion related to medical and cure of coronavirus. And, the later discusses about domestic and international politics. There are numerous rumors regarding the coronavirus's origin and spread. Some claim that this virus was produced in a laboratory in China. In contrast, others say that it originated in animals such as bats. Lots of countries have placed the blame for this virus squarely on China's shoulders. As a result, it has become involved in international political disputes. That is why we have classified these two topics as part of the *Politics* sub-category.

**Social Related Topics.** We found one sub-category and three topics under this category. The sub-category is *Education* (15.60%). It is divided into two topics: *Scientific Awareness* and *Religious Awareness*. This sub-category discusses information about Covid from the perspectives of science and religion, respectively. Another topic covers racial vilification in connection with Covid.

> *Findings* 6: Following LDA and topic labeling, the fake posts are organized into ten topics and three categories. A sizable part of misinformation falls under the *Belief* category, which includes misinformation about religion and personal belief. Numerous individuals use religion as a tool for propagating lies. *System* is the second largest category. It is divided into five topics and is concerned with medical issues, the cure for coronavirus, and the politics around it. The third category is *Social*, which contains information about Covid from a religious and scientific perspective. It also includes racial vilification.

# 6 HOW DO PEOPLE REACT ACROSS DIFFERENT TOPICS OF FAKE NEWS? (RQ4)

## 6.1 Motivation

Following an analysis of the topics, we study how people react to them. These will help us in determining how people feel about and approach particular issues. Additionally, we can decide which topics are more popular with the public and which are less popular. This will assist others in evaluating the level of acceptance and response to various issues in people's life.

## 6.2 Approach

Facebook made reactions global on February 24, 2016 [35]. Along with the *Like* button, reactions enabled users to interact with Facebook posts in a simplistic way. With a single click, users could share apparent emotional responses via these emoji-like icons (*Love*, *Haha*, *Wow*, *Sad*, and *Angry*). They are mutually exclusive, providing nuance that the rather imprecisely used Like button lacked [40].

Online user comments are probably the most researched and debated form of participation [118]. While leaving comments below news articles or social media posts, commenters express cognitive claims (e.g., promoting their experiences and contributing their views) or cope with emotions triggered by online content (e.g., express anger) [96]. Users can also interact with online content via social media (e.g., Facebook *Like*). Such *Click speech* is understood as a form (or even a substitute) of online conversation [72].

We categorize all reactions except the *Like* reaction into three categories. They are: (i) Positive, (ii) Negative, and (iii) Haha. The *Positive* group includes the *Love*, *Wow*, and *Care* reactions. *Sad* and *Angry* reactions have been integrated into the *Negative* group. Eventually, the *Haha* group consists only of *Haha* reaction. After categorizing in this way, we order the posts within every category by their total interaction count. Following this categorization, we sort the sequence of posts for every category in descending order of the cumulative number of reactions within that category. For instance, in the *Positive* category, we calculate the total amount of *Love*, *Care*, and *Wow* reactions for each article. This way, we can determine the total number of positive responses to each post. Then, we sort all of the posts in decreasing order of this number of positive responses. We follow a similar strategy for the *Negative* and *Haha* groups as well. Now we have distinct orders of posts for each category. In this part, we will utilize these data to demonstrate how individuals react differently to various topics.

## 6.3 Results

We first discuss the distribution of posts across all topics. Then we explain how individuals respond to these topics and some comments extracted from the comment sections.

**Distribution of Posts.** We conducted a demographic analysis of the frequency of posts with respect to topic from Bangladesh, India, and other nations and present that in Figure 11. In Bangladesh, the majority of posts concern Vaccination and Religious matters, which is also true for India and other countries. However, the number of posts on *Religious Rituals* in Bangladesh is substantially more than that in India. On the other hand, in terms of religious awareness and superstitious customs, India and Other countries both surpass Bangladesh. Other countries have a sparse number of postings about the origin and dissemination of the Coronavirus, whereas Bangladesh has a sizable quantity of posts on these topics.

> *Findings* 7: In all regions, more than half of fake news is about vaccination. In India and Bangladesh, religious posts on Facebook are highly prevalent. On the other hand, the propagation and origin of coronavirus receives the least attention in misinformation.

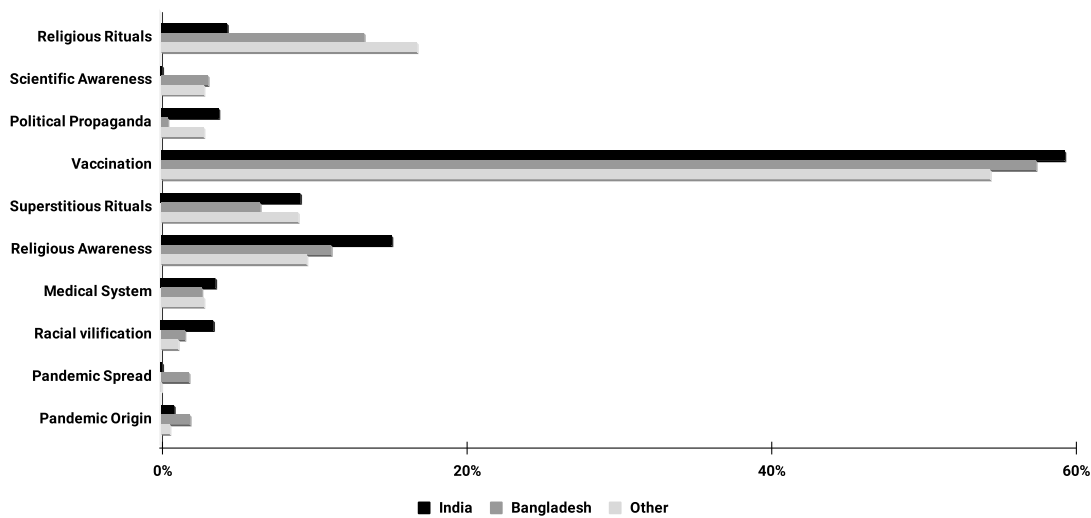

Figure 11: Distribution of post with respect to topic

**Individuals' Responses to Various Topics.** In Figure 12, we present an assessment of how people react to various topics by analyzing the responses to posts. In this scenario, we use the three previously mentioned reactions: *Positive*, *Negative* and *Haha*. It appears as though people behave differently on different topics, which is sometimes influenced by the situation they are in. For example, it is commonly acknowledged that the coronavirus originated in China. As a result, countless posts on Facebook accuse and criticize them racially. Some posts contain made-up stories that lack an original source, and hence are labeled as false. Regrettably, it appears that some individuals believe in these fraudulent news and respond positively to it. Thus they are getting involved in this racial denigration. Additionally, people appear to be responding positively to religious opinion related posts as they seek consolation in religion during the difficult situation.

On the other hand, political propaganda generates the greatest negative responses of any topic. In this difficult time, people do not appear to be enjoying this hoaxes and are reacting negatively to it. Posts on medical system and superstitious rituals are receiving most Haha reactions. Individuals in this modern era hold far fewer superstitions than they did in the past. The vast majority of these views have been eventually proven to be scientifically inaccurate. As a result, they face much mockery. This also holds for false scientific evidence. Due to the easy access to modern technology, people can verify anything much more swiftly than before. For example, there is a post saying,

> "Coronavirus will be destroyed by solar eclipse, the earth will be safe! The Indian scientist explained this with an argument. During a solar eclipse, the earth's atmosphere undergoes chemical changes, which in turn gives rise to the corona virus. And during the next solar eclipse, the existence of corona from this earth will disappear."

The same is true with conspiracy theories too. Numerous theories are available on the internet without any valid sources. These types of posts are also becoming victims of criticism.

In terms of positive reaction, posts about religious rituals receive the most comments, while those about scientific awareness receive the fewest. In terms of negative reaction, vaccination generates a lot of discussion and also generates a lot of haha reactions. A political official in India made a statement on the cure for the Covid virus, which went viral at the time and provoked a lot of laughter from the internet.

> "If you sit in the mud and play conch shells, you will not be infected by Covid."

Additionally, several newspapers covered this story [12]. There was another post that became the target of much ridicule, stating "Crowds are flocking to Gujarat's cowsheds to get rid of corona by mixing cow dung and cow dung.". Vaccination-related posts are shared more frequently than any other type of content.

> *Findings* 8: Political fake news draws more negative responses than any other type of news. However, people appear to be more responsive to misleading information related to religion than to any different topic.

**Exploring Comment Sections.** We chose the top 40-50 posts from each category and read their comment sections. We extracted one or two comments with the highest number of interactions from the comment section and displayed them in the table 10 and 11. Additionally, we have explained why this post should be classified as fake.

We came across some posts where a large number of users left identical comments despite the post receiving a significant number of reactions. This is most prevalent in posts on religious topics.

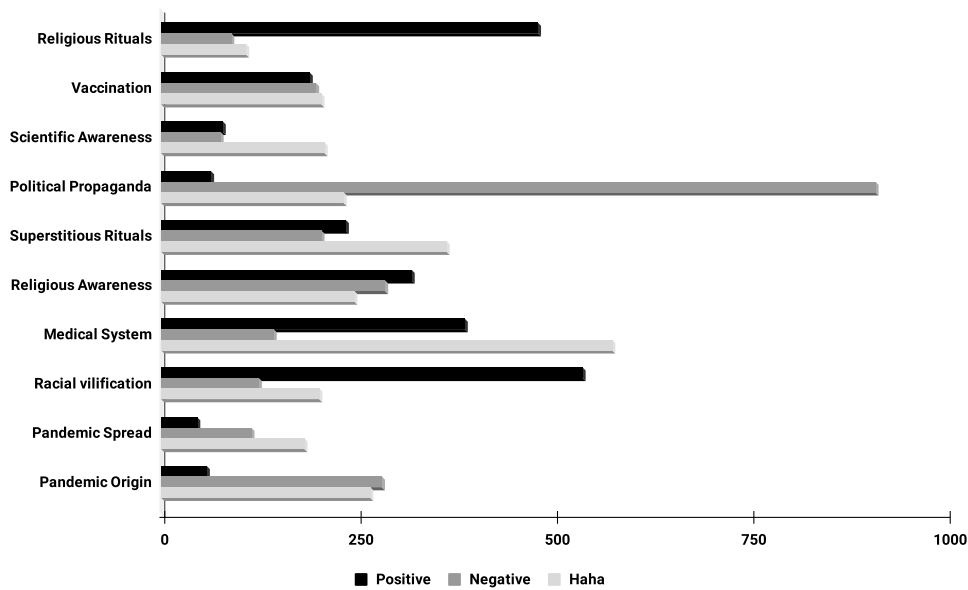

Figure 12: Average reaction on different topics

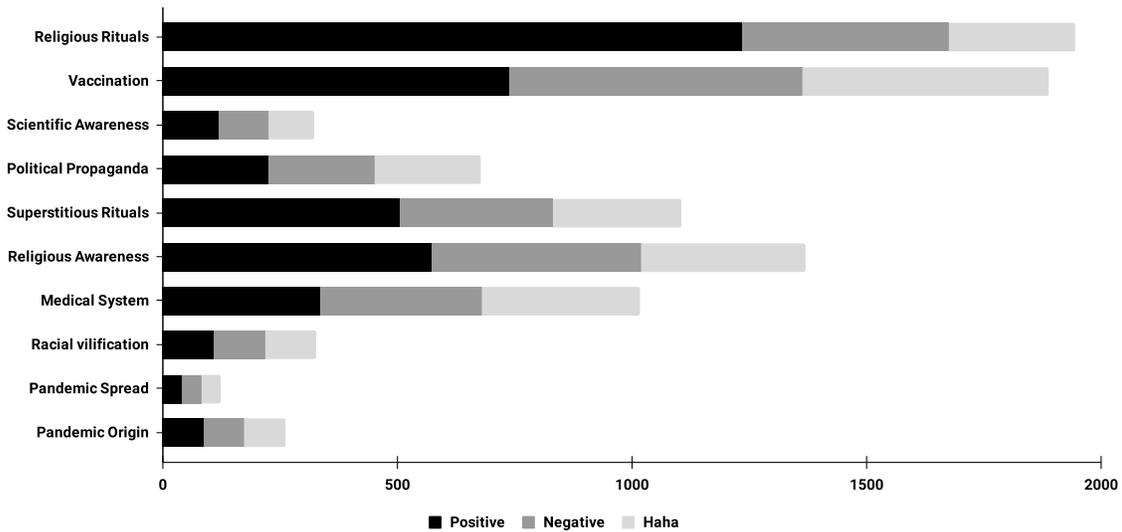

Figure 13: Average comment on different topics (Different sizes of point for each reaction group to differentiate between them.

From Figure 15, as we can see, this is a religious post and many commenters are commenting only one word.

## 7 IMPLICATIONS OF FINDINGS

Four stakeholders can benefit from our findings: (1) **Health Agencies** to design and disseminate useful guidelines for the mass, (2) **Religious Institutions** to guide people down the right path, (3) **Government Agencies and International Communities** to keep people informed and maintain peace and (4) **Common People** benefiting from the systems derived from our work.

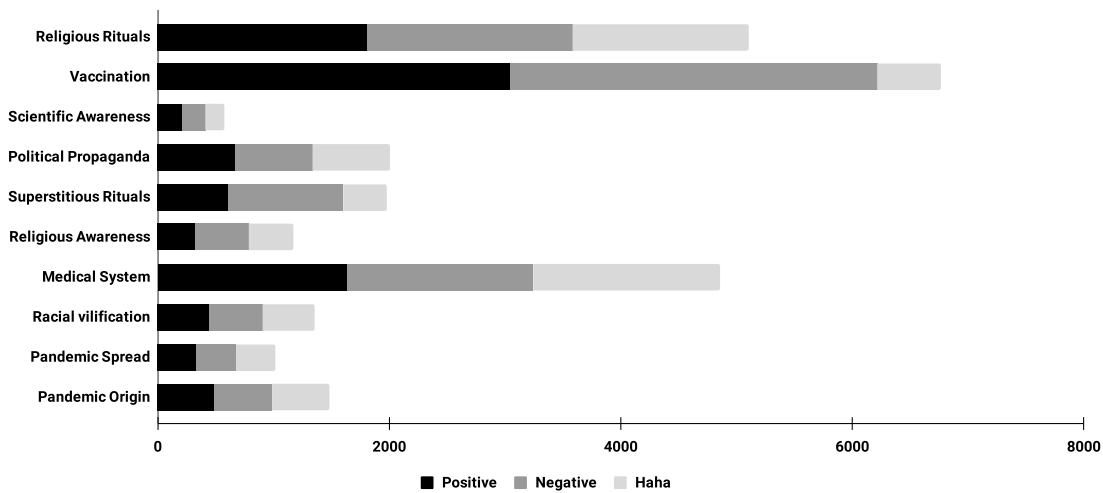

Figure 14: Average share on different topics (Different sizes of point for each reaction group to differentiate between them.)

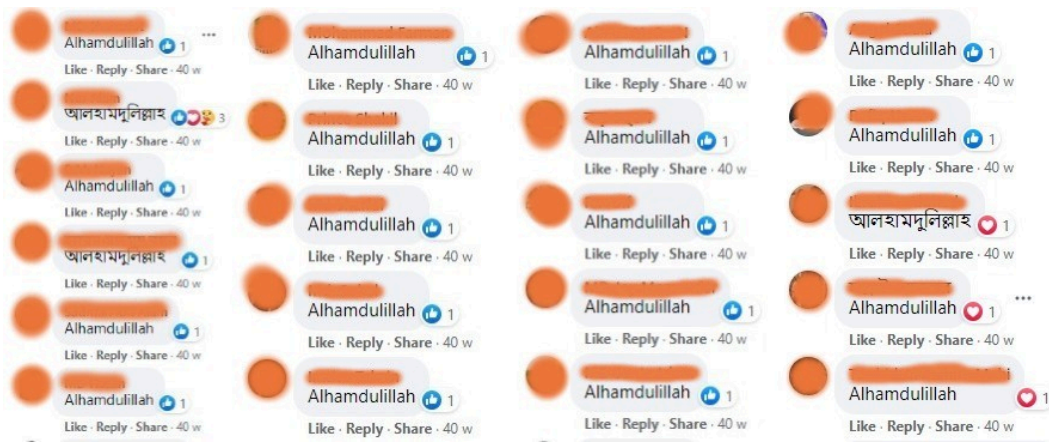

Figure 15: Identical Comments on a Post((All the orange circles and lines in the figure were post-edited for anonymization.)

Each topic we found in Section 5 is depicted in Figure 16 in terms of popularity vs the proportion of negative reactions. Here, popularity indicates the traction a post gained on social media. The measure of traction is numerically represented with the number of shares, comments and *Like Reaction*. As per Section 6.3, *Like* is not listed among the *Positive Reactions*. Together with comments and shares, *Like Reaction* comprises a judgment irrespective (whether positive or negative) form of interest a Facebook post has inspired. Negative reaction is defined in the same way as in Section 6.3. The size of each bubble represents the total number of posts regarding the corresponding topic.

Since ours is an empirical study, we delineate our observations in the following and substantiate our recommendations on how the stakeholders can utilize our findings about the emerging themes in Figure 16.

**Health Agencies.** Since this study expounds on how Facebook users behave towards fake news during a pandemic, our work has significant implications for health agencies and relevant research organizations such as IECDR, CDC, WHO. A considerable number of posts are about vaccination (around 23.74%) as we see from Figure 10. Conversely, misinformation about the medical system accounts for only 1.55% of the fake posts. However, fake news regarding vaccination attracted the most negative reactions (about 49%) as opposed to the largely accepted myths about medicines and cures, as Figure 16 tells us.

For instance, this following *post* experienced wide acceptance. This post received 53 thousand likes, 12 thousand shares and 1.5 thousand comments. Whereas, it incurred fewer than a hundred negative reactions.

> *A drug called 'Avigan' has succeeded in a trial to prevent coronavirus. A group of Japanese researchers claim*

Table 10: Posts with most positive reaction

| Topics | Post(s) | Selected Comment(s) | Verdict |
|---|---|---|---|
| **Religious Ritual** | "Worship of 1023-year-old Mahakali Mrinmayi Goddess begins in Bishnupur from the belief that she has brought the virus here and only she can get rid of it." | "Who says there will be no profit? Remember that the world survives on this belief. Otherwise, when would this world be destroyed!" | One person supports it, while another protests and calls it fake news. |
| | | "There is no benefit in doing all this, nothing happens, all fake." | |
| | "Corona will end on May 12 !!! Wondering? Let's match what is said in the hadith! On that day, you will completely wipe out the plague from the world; take it out of the world! Take it away from the world. On the day when they rise to the earth, all plagues will be removed from the earth. (Musnad Ahmad, Hadith No. 747)" | "If you believe in it, your faith will be ruined because there is a hadith in the surah stating that if the Suraiya star rise, the epidemic will be over, but May 12 does not specify that, if this virus exits after May 12, then everyone's faith in Muslims will be gone. Fear Allah, seek forgiveness from Allah, repent, inshallah, He will fix everything." | The same as before. This case has no supporting evidence, and also Covid is still active, so it's a hoax. |
| | | "Pray 5 times and repent to Allah so that Allah may forgive us all. Don't listen to what anyone is saying; put your trust in Allah, and you will see that everything will be fine." | |
| **Political Propaganda** | "Corona is spreading across the state for the BJP government." | "This is the beginning of blaming leaders, this time, only ordinary people will die. If the leaders get infected by Corona, they will be provided a seat in the hospital, but ordinary people will not." | BJP, a political party in India, is blamed for spreading Covid across the nation. Most of the people in the comment section knew that this is fake and objectified it. |
| | | "This is nonsense." | |
| **Superstitious Rituals** | "If fasting can destroy virus-like cancer, then the Covid virus will also be destroyed during the month of Ramadan." | "Encourage fasting in a different way, not this way." | Some people apply religious belief to regular activities. A few people support it, while others protest it. |
| | | "In sha'Allah" | |
| **Religious Awareness** | "To stop the spread of coronavirus, quarantine is the only option. It must be accepted. But it was the Prophet Muhammad who first mentioned it. So says a US researcher." | "Subhanallah, O Messenger of Allah, show me the way to get rid of this evil. Amen." | No credible US researcher has stated this. This is clearly a made-up story. But it's hard to verify on Facebook, so some people believe it. |
| | | "To my knowledge, no American Scientist has said this." | |
| | "The name of the coronavirus drug is Q7. Here, Q is for Qur'an, and 7 is for seven verses of Surah Fatiha." | "Don't write something made up. You will not get rid of sin instead of virtue." | This is obviously fake news, which people seem to accept. |
| | | "Did you make this up or is this the official description?" | |

that a patient can recover in just seven to nine days by taking this drug.

The following viral fake post about a vaccination equivalency has not attained considerable positive feedback.

The search for a vaccine for coronavirus is already underway in India. In the meantime, the Congress councilor of Karnataka went viral overnight by talking about a sensational method to prevent coronavirus! The Congress councilor is known as a social worker of Karnataka.

Table 11: Posts with most negative reaction

| Topics | Post(s) | Selected Comment(s) | Verdict |
|---|---|---|---|
| **Religious Ritual** | "If you work for 8 hours in garments, you will not be infected by Covid. But if you pray for 6 minutes in a mosque, you will be infected. Strange world." | "If the prayers are offered at home, Allah will accept them, but our poor people will face food and economic problems if the garments are closed for a long time." | During the initial phase of Covid's dissemination, the government prohibited all citizens from visiting mosques and advised them to pray at home. This post mocks this notion. The majority of people appear to be opposed to the idea expressed in this post. |
| | | "Is everyone in agreement that the government is refusing to travel to a large number of locations? How many of those actually go to the mosque and pray five times a day? Bengalis are not useless; we are some people who are unaware of their own mistakes." | |
| **Scientific Awareness** | "If he was not promoted to the post of professor, an associate professor threatened to disseminate covid." | "Good initiative" | Someone wants to frame an assistant professor at a prestigious university. Few people believe it, as evidenced by the reaction count. However, others oppose it through mocking. |
| | | "Bengalis want to deceive others by taking advantage of the epidemic." | |
| **Superstitious Ritual** | "Five minutes after birth, the baby is saying the way to get rid of the corona." | "I am drawing the attention of the administration. Rumors have terrorized ordinary people! Let legal action be taken against them. Everyone Raises their voice. Build a rumor-free country." | This is not scientifically possible. People got it and voiced against it. |
| | | "The shepherd's event will occur one day. The tiger will come, but no one will come to save." | |
| | | "I am telling you to listen to rumors. This incident is not a hundred percent true." | |

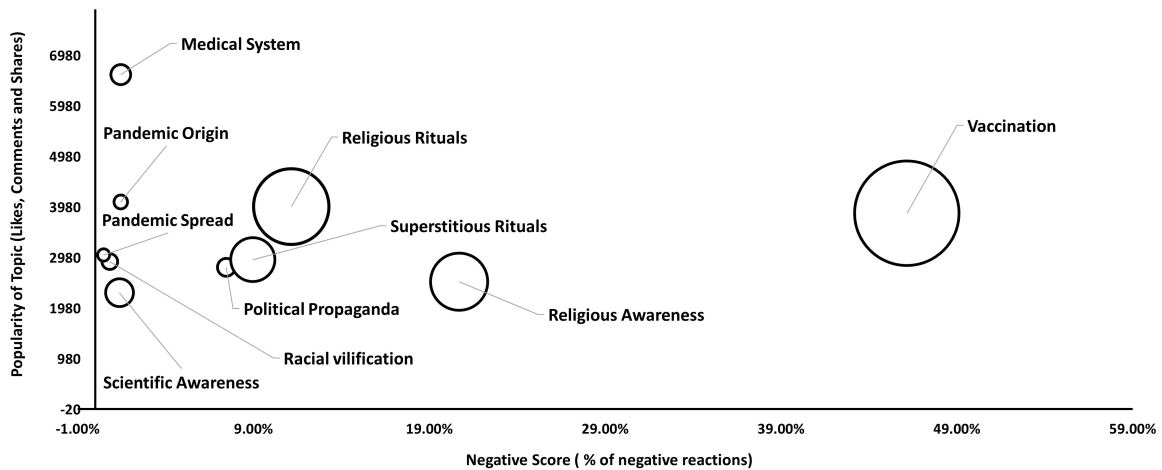

Figure 16: Tradeoff between topic popularity and negative score

*He says that 90 mg of 'Rum' should be mixed with one spoon of black pepper. Then you have to eat 2 eggs immediately to stop the effectiveness of coronavirus.*

From this observation, we can safely infer that the common mass is not easily fooled by misinformation circulating about vaccination and prevention schemes. However, people are more likely to believe fabricated information on the virus's origins, its spread, medicine, cures, and the medical system in general. Any fake news regarding Covid-19 can lead to serious health concerns. Therefore, the concerned authorities can learn why people are less prone to believe fake news regarding vaccines and devise ways to provide the mass with legitimate information effectively.

**Religious Institution.** Religious institution denotes any organization that is in charge of maintaining and regulating the faith of the citizens. Prayer houses of various religions, governments, or non-government entities affiliated with such tasks are religious institutions. Religion is a matter of faith. We see from Figure 16 that fake information veiled by religious rituals accounts for a significant portion of the total fake posts but they received about 10% of the total negative reactions. This is an alarming trend because it is easy to manipulate and misguide people using religion as a tool. A quintessential example would be the following *post* that engendered significant number of *Likes* (almost 3 thousand) and comments.

> *Coronavirus will not transmit during the month of Ramadan: WHO*

Perpetrators can use the gullibility of the humans to their advantage. Hence, it falls upon the religious institutions to debunk myths such as this floating around social media and guide people in the right way.

**Government and International Community.** Fake information regarding political propaganda, racial vilification, superstitious rituals together received below 20% (refer to Figure 16) of the total negative reactions proving people are likely to be hoodwinked by them. On the other hand, each of these topics occupy about $\frac{1}{3}$ of the distance along the popularity axis, as depicted in Figure 16. Let's take a look at the post below.

> *The coronavirus was spread from one of the laboratories in Wuhan, and the United States leaked a secret cable to Foreign Office containing talks with US embassy officials in China to reinforce the claim. The Trump administration has been claiming since April that the coronavirus has spread from its laboratory, not from bats or pangolins. The same thing was heard from President Donald Trump in May. "I have the documents", Trump said. "I'm sure it spread from Wuhan's lab. But I can't say how I got it. I don't think so."*

Ignorance in concert with misinformation such as the above can result in commotion, even racial violence. For example, a student was assaulted in London for his race's alleged ties with spreading coronavirus [26]. Governments can monitor the misinformation prevailing on social media and prevent such disorders from occurring with the help of law enforcement agencies. International communities also have responsibilities to raise awareness against racial prejudice stemming from such fake news.

**Common people.** General people suffer the most due to false information. Covid-19 being an unprecedented situation, people may be at a loss as to what should be the appropriate step. Fake posts about scientific awareness comprise a substantial portion of fake posts. However, posts as such received insignificant proportion (less than 5%) of disapproval as evident from Figure 16. Posts such as the following can lead people astray.

> *Meanwhile, a Chinese expert has assured everyone that the coronavirus is* 100% *vanquished as a result of inhaling steam. The coronavirus cannot withstand hot water vapor, even if the virus infects the nose, throat, or lungs.*

Therefore, our analysis can help people, in general, distinguish correct scientific facts from simply made-up information. Moreover, if any fake news detection system is devised based on our work and made available for public use, the citizens can stay informed and not be misled.

## 8 THREATS TO VALIDITY

To ensure reproducibility of the analyses presented in this paper, we share our benchmark dataset in an online repository. We discuss threats to the validity of our study following common guidelines for empirical studies [110].

**External Validity** threats concern about the generalizability of our findings. We concentrated on Facebook, the world's largest and most popular social networking platform. Our findings, however, may not be applicable to other social networking platforms. The accuracy of our evaluation of fake news depends on our ability to correctly identify and label the fake Bengali posts we analyzed. At times, determining the validity of information based just on its content can be rather challenging. Each news report needs to be thoroughly investigated by comparison to other news stories on the internet. As Covid-19 is a recent phenomenon, lots of information about this virus are still under study. As a result, information currently believed to be erroneous may be proven to be right in the future. Our study considered a piece of news misleading based on the information that has come to light so far.

**Internal Validity** is compromised due to experimental bias and analysis errors. We manually labeled the topics in our study, in particular. To exclude any bias in the labeling, three distinct authors labeled the topics independently. They then evaluated them with the help of two other authors. All of the authors discussed it and worked out any disputes. As a result, we believe we have minimized labeling bias.

**Construct Validity** threats relate to probable errors that may occur during data extraction. As described in Section 3.3.2, we discarded posts that met particular criteria while constructing our dataset. For instance, we excluded duplicate, irrelevant, and unverifiable posts. Additionally, we removed posts that addressed religious faith or that contradicted any scientific fact. Furthermore, posts concerning the local death toll and infection rate were omitted from our dataset. As a result, we did not train our models on posts that were either too sensitive or irrelevant to include in the dataset. Then, from March 2020 to June 2021, we created a list of all the fake posts predicted by our model, totaling 6012 counterfeit posts. These 6012 posts may contain news that was censored out of our dataset initially.

## 9 CONCLUSION

During troublesome times such as the ongoing Covid-19 pandemic, misinformation has run rampant and exacerbated the state of chaos and confusion in society. In this paper, we have studied the detection of Covid-related fake news in Bengali Facebook posts. Firstly, we used CrowdTangle to extract Bengali posts about Covid and built a benchmark dataset containing such posts. Secondly, We investigated with three pre-trained advanced language models (BERT, RoBERTa and DistilBERT) to ascertain their efficacy with regards to fake news detection on our dataset. We found that BERT stands out as the best model with an F1-score of 0.97. Thirdly, we employed our best-performing model to filter out all the fake news from Covid-related posts between March 2020 and June 2021. We further studied several characteristics of those fake posts. These properties include origin, types of posts (Link, Status and Live Video), how people respond to fake posts, how fake information fares in the eyes of Facebook users, etc. Finally, we applied topic modeling to discern the general topics among the circulating misinformation. We identified ten topics grouped into three categories (System, Belief and Social). Our future work will investigate the impact of the fake news on Bengali speaking people by conducting surveys and within the contextual setup of diverse life scenarios (e.g., willingness to vaccinate among the religious vs non-religious people).